\documentclass[twoside,leqno,twocolumn]{article}

\usepackage[letterpaper]{geometry}

\usepackage[numbers]{natbib}

\usepackage{ltexpprt}
\usepackage{graphicx}
\usepackage{times}
\usepackage{soul}
\usepackage{url}
\usepackage[utf8]{inputenc}
\usepackage[small]{caption}
\usepackage{graphicx}
\usepackage{amsmath}
\usepackage{booktabs}
\usepackage{algorithm}
\usepackage{algorithmic}
\usepackage{xcolor}
\usepackage{booktabs}
\usepackage{multirow}
\usepackage{amssymb}
\makeatletter
\def\hlinewd#1{
\noalign{\ifnum0=`}\fi\hrule \@height #1 
\futurelet\reserved@a\@xhline}
\makeatother
\usepackage{diagbox}
\newcommand{\remove}[1]{}

\newenvironment{talign}
 {\align}
 {\endalign}
\newenvironment{talign*}
 {\csname align*\endcsname}
 {\endalign}

\begin{document}

\newcommand\relatedversion{}


\title{\Large Modality-aware Transformer for Financial Time series Forecasting }
\author{Hajar Emami \thanks{E-mail:hajar.emami@ibm.com}\qquad
 Xuan-Hong Dang\qquad
 Yousaf Shah\qquad
 Petros Zerfos
\\IBM T.J. Watson Research Center}

\date{}

\maketitle

\begin{abstract} \small\baselineskip=9pt 
Time series forecasting presents a significant challenge, particularly when its accuracy relies on external data sources rather than solely on historical values. This issue is prevalent in the financial sector, where the future behavior of time series is often intricately linked to information derived from various textual reports and a multitude of economic indicators. In practice, the key challenge lies in constructing a reliable time series forecasting model capable of harnessing data from diverse sources and extracting valuable insights to predict the target time series accurately. In this work, we tackle this challenging problem and introduce a novel multimodal transformer-based model named the \textit{Modality-aware Transformer}. Our model excels in exploring the power of both categorical text and numerical timeseries to forecast the target time series effectively while providing insights through its neural attention mechanism. To achieve this, we develop feature-level attention layers that encourage the model to focus on the most relevant features within each data modality. By incorporating the proposed feature-level attention, we develop a novel Intra-modal multi-head attention (MHA),
Inter-modal MHA and Target-modal MHA in a way that both
feature and temporal attentions are incorporated in MHAs. This enables the MHAs to generate temporal attentions with consideration of modality and feature importance which leads to more informative embeddings. The proposed modality-aware structure enables the model to effectively exploit information within each modality as well as foster cross-modal understanding. Our extensive experiments on financial datasets demonstrate that Modality-aware Transformer outperforms existing methods, offering a novel and practical solution to the complex challenges of multi-modal financial time series forecasting.
\end{abstract}


\section{Introduction}
Prediction models have been widely used for generating information to use in various decision making applications, e.g., retail demand prediction for managing the inventory \cite{riemer2016correcting}, user action prediction in advertisement applications \cite{yang2017local}, disease propagation prediction \cite{matsubara2014funnel} in clinical practice, and financial market movement prediction for portfolio management \cite{de2018advances}.
Recent deep forecasting models, especially the transformer-based models \cite{zhou2021informer,kitaev2020reformer,niu2023time} have achieved great success in time series forecasting. However, existing methods primarily operate within a single input modality framework, relying solely on historical time series data to predict future values.
Forecasting behaviors of time series is a challenging task particularly when its performance is highly influenced by external data sources rather than its own past values. One such valuable external resource is textual information, which often contains signals that aid in forecasting the future behavior of certain time series. Examples of this synergy between text data and time series forecasting are widespread in the financial domain. In this domain, large amounts of textual and time series data are generated daily, offering the potential to enhance the prediction of various financial time series. For instance, statements released by the Federal Reserve (FED) and Beige Books, which provide insights into economic conditions, contain crucial information that can significantly influence the forecasting of U.S. interest rates across different maturities. Interest rates are vital indicators of economic growth and are integral to the financial decision-making process. Economic and market events can trigger fluctuations in interest rates, making them essential variables to monitor and predict \cite{nabil2020survey}.

Forecasting the future movements of interest rates is one of the most challenging problems in the finance area. The nonlinear and dynamic nature of interest rates complicate this prediction task. 
The need to incorporate multiple data sources (under different data formats) and extract useful signals related to the financial time series are required to ensure the success of a financial time series forecasting model. 
One major challenge of existing transformer-based models for this task often lies in constructing a powerful model that can maximally exploits the information embedded in data from the textual modality along with various indices of the economy from the numerical time series modality.
Integrating information from multiple data sources in multimodal learning with transformers has been studied in different applications \cite{shi2022learning,daiya2021stock,zhan2021product1m}. 
These studies combine and treat different modalities as one single data modality processed by conventional transformer blocks in either encoder or decoder or both. Hence, they are limited in exploiting the cross-modal interactions between different modalities at all levels of the encoder and decoder. 
In \cite{xu2023multimodal,zhan2021product1m}, multi-modal transformer models require an additional conventional transformer to model the global context, which may cause computational and memory limitations.

To overcome the above research issues, we propose a novel Modality-aware transformer-based model (MAT) that can effectively utilize multiple data modalities for predicting a target time series along with explanation through its neural attention mechanism. 
We introduce feature-level attention layers to extract the most relevant features in each modality. 
That is, Modality-aware Transformer gradually assigns larger attention weights to the most informative text topics and time series indices, while pays less attentions on little or no relevant ones. 
By incorporating the proposed feature-level attention, we develop a novel Intra-modal Multi-Head Attention (MHA) and Inter-modal MHA in our modality-aware encoder in a way that both feature and temporal attentions are incorporated in MHAs. This enables the MHAs to generate temporal attentions with consideration of modality and feature importance which leads to more informative embeddings. 
We also develop a novel Target-modal cross MHA fused by feature-level attentions in the decoder which enables the model to discover the interactions between the target series and learnt patterns from two input data modalities toward decoding (forecasting) behavior of the target time series in the next timepoint(s).

It is important to note that, compared to existing transformer models, our proposed model consists of novel transformer blocks with intra-modal MHA, inter-modal MHA and target-modal MHA to extract both temporal and feature attentions at all levels of the encoder and decoder, being more effective in multimodal learning. It hence does not need an additional conventional transformer to predict future behaviors of the target time series.
Furthermore, the proposed modality-aware structure naturally provides us the great flexibility to extend MAT to the case associated with a flexible alignment at the timestamp level between the two data modalities. Intuitively, the presented fashion enables the proposed model to deal with two input data sequences of different lengths for all potential timestamp alignments. The major contribution of our work is summarized as follows:
\begin{itemize}
    \item We propose Modality-aware Transformer to successfully enhance the prediction capacity in financial time series forecasting, which validates the transformer-based model’s potential value to capture information embedded in financial textual reports and time series modalities in order to predict future values of time series.
    \item We develop feature-level attention layers to help the model focus more on the most relevant features in each modality. This attention provides a natural interpretation of relevant features learnt from each data modality. 
    \item We develop novel Intra-modal MHA, Inter-modal MHA and Target-modal MHA fused by the feature-level attentions
    to generate temporal attentions with consideration of modality and feature importance which leads to more informative embeddings.
    The developed MHAs enables the model to effectively exploit information within each modality as well as across modalities. 
    \item Through extensive experiments, we show that, MAT significantly out-performs other state-of-the-art time series forecasting methods over FED statements/BeigeBooks and U.S. interest rates dataset. 
\end{itemize}

The remainder of this paper is organized as follows: Section
2 contains a brief review of the literature surrounding time series forecasting and multimodal learning. In Section 3, we present the problem definition and introduce our Modality-aware Transformer in detail. The experimental setup is presented in Section 4. The experimental results and evaluation are discussed in Section 5.
Finally, we conclude in Section 6.

\section{Related Work}
\subsection{Time series Forecasting}
Many classical approaches have been developed to solve time series forecasting problems, such as Auto Regressive Integrated Moving Average (ARIMA) \cite{box2015time} or exponential smoothing \cite{hyndman2008forecasting}. However, they are limited in predicting complex time series data.
Recently, deep learning methods have been employed in time series forecasting problem, including convolutional neural network (CNN) \cite{munir2018deepant}, 
including recurrent neural network (RNN) \cite{salinas2020deepar}, and Generative Adversarial Nets (GANs) \cite{taymouri2021deep}. 
There are recent interests of using Transformers in time series problems, leading to great progress in various time series applications. The self-attention mechanism in transformers improves the performance by enabling the model in capturing long-term dependencies in sequential data \cite{kitaev2020reformer,wu2021autoformer}.
Informer \cite{zhou2021informer} proposed by Zhou et al. for long term time series forecasting, designs a generative style decoder to produce long term forecasting directly to avoid accumulative error in using one forward step prediction. 
Tong et al. combined the Transformer and GAN and proposed the Probabilistic Decomposition Transformer \cite{tong2023probabilistic}, for time series forecasting. 
Their proposed mechanism utilizes the forecasting results of Transformer as conditional information for the generative model, performing sequence-level forecastsing.
However, the aforementioned techniques treat different modalities as one single modality, hence are limited in exploiting complementary information and lack the capability of providing interpretation based on information embedded in the text modality and time series modality. 
Furthermore, they have constrains on the time-step synchronization between different modalities.

\subsection{Multimodal Learning}
Multimodal learning has brought great interests recently due to its significant advances in many applications. 
Incorporating time series and textual corpora as input data sources is becoming a promising approach, especially in the financial industry. 
Gunduz et al. \cite{gunduz2015borsa}, proposed an approach that forms the textual feature vector based on BoW representation and combines with the daily, weekly and monthly moving average features computed from open, close price and the volume of stocks. The Naive Bayes classifier has been used to predict the stock movement of the next day.
In \cite{weng2017stock}, the daily counts of Google News along with a number of technical indicators are incorporated to predict the movement of stocks under concerns. 
The work presented by Akita et al. \cite{akita2016deep}, handles the (company) stock data in its natural form of time series and subsequently being merged with the vector representation of textual news prior to making the market prediction. 
Adämmer et al. \cite{adammer2020forecasting}, grouped news articles from the NY Times and Washington Posts 
into topics using the correlated topic modeling method by treating each article as a vector of word accounts based on bag of words. The derived topics are manually selected by domain experts and the model is not able to handle external time series.

In \cite{xu2023multimodal} multimodal learning with transformers are classified into six categories: “Early Summation" \cite{gavrilyuk2020actor} and “Early Concatenation” \cite{shi2022learning} studies that generally have a single stream structure in which token embeddings from multiple modalities are summed or concatenated and then considered as a single data modality processed by the conventional transformer. 
“One-stream to multi-stream” \cite{lin2020interbert} and “Multi-stream to one-stream” \cite{daiya2021stock} studies that have hybrid-stream structure (a combination of single-stream and multi-stream). Similar to the first two categories, these studies also treat different modalities as one single modality in either encoder or decoder. 
All aforementioned studies employ conventional transformers which have only self-MHA and thus are limited in exploiting the cross-modal interactions between different modalities, and are also limited in dealing with asynchronous data modalities such as categorical text and numerical time series, with different data length and data sampled frequencies. 
These studies also lack the capability of providing interpretation based on information embedded in different modalities.
In “Cross-Attention” studies \cite{murahari2020large}, transformer blocks only include simple cross-MHAs with no self-MHA which can lead to global context loss. “Cross-Attention to Concatenation” studies \cite{zhan2021product1m} alleviate this issue, in which two streams of cross-attention are further combined and treated as a single data modality processed by an additional conventional transformer to model the global context. 
Thus, their decoder is limited in discovering the correlated patterns between the target time series and each data modality at different timesteps.
Furthermore, requiring an additional conventional transformer, may cause computational and memory limitations for these studies.

\begin{figure*}[htb]
\centering
    \includegraphics[width=0.8\linewidth]{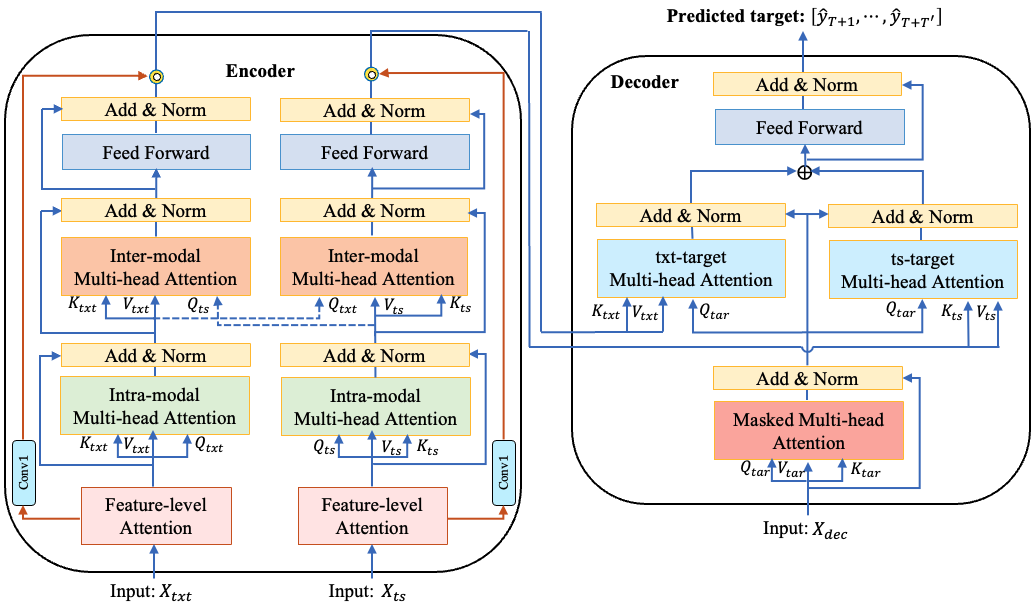}
    \caption{The Modality-aware Transformer (MAT) architecture introduces innovative components, including Intra-modal MHA, Inter-modal MHA, and Target-modal MHA, all fused through feature-level attentions. This fusion results in enhanced temporal attentions, considering both modality and feature significance, thereby generating more informative embeddings. MAT's modality-aware design empowers it to efficiently utilize information within and across modalities.}
    \vspace*{-0.3cm}
\label{fig:MAT}
\end{figure*}

\section{Methodology}

\subsection{Problem Definition}
Multi-step-ahead time series forecasting problem can be formulated as predicting the target variable, $\hat{y}$, for $T^\prime$ steps in the future given the past $T$ observations of the input sequences and the target variable. The model can be described as $\hat{Y}=F(X)$ where $\hat{Y}=(\hat{y}_{T+1}, \hat{y}_{T+2},...,\hat{y}_{T+T^\prime}) \in \mathbb{R}^{T^\prime}$ represents a sequence of target variable $\hat{y} \in \mathbb{R}$ for the future $T^\prime$ steps, $X=\{x_t \}_{t \in T}$ is an ordered set (i.e., lookback window) of past $T$ observations of $q$ inputs and one target output and $x_t = (x_{t}^1,x_{t}^2,...,x_{t}^q,y_t) \in \mathbb{R}^{q+1}$ is the observation at a specific time $t \in T \subseteq \mathbb{Z}^+$. 
The model’s output $\hat{Y}$ is the predictions of the target variable in the prediction horizon $T^\prime$. In multimodal time series forecasting, $X$ can be from multiple data modalities.
The goal of the proposed approach is to effectively utilize data from both text modalities $X_{txt}$ , and data from time series $X_{ts}$ for predicting target time series.

\subsection{Modality-aware Transformer}
Figure~\ref{fig:MAT} shows  the  architecture of  the proposed  MAT  model.
The proposed model follows the encoder-decoder architecture of the transformer-based models \cite{vaswani2017attention}. 
To maximally exploiting information embedded in different modalities, we introduce the modality-aware structure to the Transformer architecture which helps the model to effectively exploit information within each modality as well as across modalities to better forecasting future behavior of a related target time series.
We extend the transformer model with additional feature-level attention layers to focus more on the most relevant features in each modality. 
By incorporating the proposed feature-level attention, we develop a novel Intra-modal MHA and Inter-modal MHA in our modality-aware structure in a way that both feature and temporal attentions are incorporated into MHAs. We extend the modality-aware mechanism into the decoder to further exploit the dependencies between the target time series and each data modality effectively, leading to forecasting improvement. Different from the existing methods, MAT can generate different weights for the same timesteps yet in different modalities.

\subsubsection{Feature-level Attention}
As illustrated in Figure~\ref{fig:MAT}, data from each modality are fed to the feature-level attention layer to extract the most informative features for each modality. This soft attention layer places the attention weights over the input and outputs the weighted input features in a way that most informative features from different modalities can get higher attention weights if they contain predictive signals toward forecasting the target time series. The output of the feature-level attention layers are the modality feature attention weights and the weighted input modality $X_{mod_i}^{att}$:
\vspace*{-0.2cm}
\begin{align}
    X_{mod_i}^{att} = \operatorname{softmax}\left(conv1(X_{mod_i})\right) \circ X_{mod_i}
\end{align}

\noindent where $X_{mod_i}$ is the input data sample from $i$-th modality, and $\circ$ is element-wise multiplication.
This feature-level attention provides a natural interpretation of relevant features learnt from each input data modality. 

By incorporating the proposed feature-level attention, we develop a novel intra-modal MHA and inter-modal MHA in a way that both feature and temporal attentions are incorporated in MHAs. This enables the MHAs to generate temporal attentions with consideration of modality and feature importance which leads to more informative embeddings. 
We also incorporate the proposed feature-level attention into our decoder and develop a novel target-modal attention mechanism fused by feature-level attentions which aims at discovering the correlated patterns between the target time series and each input data modality, toward decoding behavior of the target time series in the next timepoint(s). Feature-level attention layers provide important information for understanding the association between different features and the target time series.

\subsubsection{Modality-aware Encoder}
In the proposed MAT, the modality-aware encoder is introduced for maximally exploiting information embedded in different modalities. 
As illustrated in Figure~\ref{fig:MAT}, MAT has two separate modality streams in the encoder for extracting information embedded in different modalities. 
We develop a novel attention mechanism through intra-modal MHAs and inter-modal MHAs in the encoder. While intra-modal MHAs allow the model to attend on most important timesteps in each individual modality, inter-modal MHAs enable the model to discover the cross-correlation between different modalities.

 The intra-modal MHA in MAT transforms the weighted input into query $Q_{mod_i}$, key $K_{mod_i}$ and value $V_{mod_i}$ matrices, and then computes the scaled dot-product attention in each head. The $Attn_{mod_i}^{intra}$ are generated by applying intra-modal MHAs on encodings of different modalities:

\vspace*{-0.4cm}
\begin{talign}
    Q_{mod_i} &= Attn_{mod_i}^{feat} (X_{mod_i}) \circ W_{mod_i}^Q\\
    K_{mod_i} &= Attn_{mod_i}^{feat} (X_{mod_i}) \circ W_{mod_i}^K\\
    V_{mod_i} &= Attn_{mod_i}^{feat} (X_{mod_i}) \circ W_{mod_i}^V
\end{talign}

\begin{align}
    Attn_{mod_i}^{intra} &= \operatorname{softmax}\left(\frac{Q_{mod_i}K_{mod_i}^T}{\sqrt{{d^k_{mod_i}}}}\right)V_{mod_i}
\end{align}

\vspace*{-0.7cm}
\begin{align}
    H_p = Attn_{mod_i}^{intra}(Q_{mod_i}W_{p}^Q, K_{mod_i}W_{p}^K, V_{mod_i}W_{p}^V)
\end{align}

\noindent where 
\vspace*{-0.2cm}
\begin{talign}
    Q_{mod_i} &\in \mathbb{R}^{L_{{mod_i}}^Q\times d_{{mod_i}}^q}\\
    K_{mod_i} &\in \mathbb{R}^{L_{{mod_i}}^K\times d_{{mod_i}}^k}\\
    V_{mod_i} &\in \mathbb{R}^{L_{{mod_i}}^V\times d_{{mod_i}}^v}
\end{talign}

\noindent $L_{{mod_i}}^Q$, $L_{{mod_i}}^K$, and $L_{{mod_i}}^V$ respectively denote the lengths of queries, keys, and values in the $i$-th data modality. Likewise, $d_{{mod_i}}^q$, $d_{{mod_i}}^k$ and $d_{{mod_i}}^v$ are respectively their dimensions in this $i$-th modality.
The output of the MHA in each modality is the concatenation of the attentions in all heads:

\begin{align}
\begin{split}
    Intra\textendash modal\ MHA = Concat(H_1,...,H_h)W_{mod_i}^O
\end{split}
\end{align}

\noindent where $h$ is the number of heads in MHA and $W_{mod_i}^O$ are learnable parameters. 
Intra-modal MHA follows by layer normalization:
\begin{align}
\begin{split}
    z_{mod_i}^{intra} = LayerNorm\Big(Attn_{mod_i}^{feat} (X_{mod_i}) + \\Intra\textendash modal MHA\big(Attn_{mod_i}^{feat} (X_{mod_i})\big)\Big)
\end{split}
\end{align}

Incorporating the feature-level attention into our intra-modal MHA enables the MHAs to generate temporal attentions with consideration of modality and feature importance leads to more informative embeddings. 

While queries, keys and values are from the same modality in intra-modal MHAs, queries are exchanged between two modalities in our developed inter-modal MHA. This enables the model to discover the cross-correlation between different modalities. The $Attn_{mod_i}^{inter}$ is defined as: 

\vspace*{-0.4cm}
\begin{align}
    Attn_{mod_i}^{inter} = \operatorname{softmax}\left(\frac{Q_{mod_j}K_{mod_i}^T}{\sqrt{{d^k_{mod_i}}}}\right)V_{mod_i}
\end{align}

\vspace*{-0.4cm}
\begin{talign}
    H_p = Attn_{mod_i}^{inter}\left(Q_{mod_j}W_{p}^Q, K_{mod_i}W_{p}^K, V_{mod_i}W_{p}^V\right)
\end{talign}

\vspace*{-0.4cm}
\begin{align}
\begin{split}
    Inter\textendash modal\ MHA = Concat\big(H_1,...,H_h\big)W_{mod_i}^O
\end{split}
\end{align}

The developed inter-modal MHAs help the model to effectively deal with cases where $modality_i$ might have high correlation with some lagged orders of $modality_j$. As illustrated in Figure~\ref{fig:MAT}, the output of MHA layers will go through layer normalization and Feed-Forward Networks (FFN) with two layers and a ReLU activation function:

\vspace*{-0.4cm}
\begin{align}
\begin{split}
    z_{mod_i}^{inter} &= LayerNorm\Big(z_{mod_i}^{intra} + \\&Inter\textendash modal MHA
    \left(z_{mod_i}^{intra}, z_{mod_j}^{intra}\right)\Big)
\end{split}
\end{align}

\vspace*{-0.4cm}
\begin{align}
\begin{split}
    enc_{mod_i}^{out} = LayerNorm\left(z_{mod_i}^{inter} + FFN\left(z_{mod_i}^{inter}\right)\right)
\end{split}
\end{align}

Different from the existing methods, the proposed modality-aware structure enables the model to assign different weights for the same timestep in different modalities to capture the most informative timesteps in each modality.

\begin{figure}[tb]
\centering
    \includegraphics[width=0.8\columnwidth]{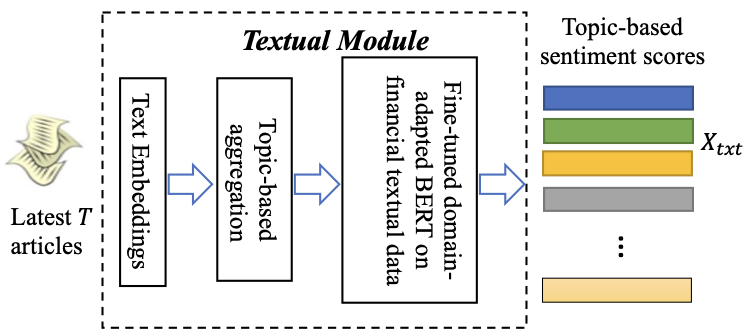}
    \caption{Textual module for representing input text reports.}
\label{fig:text}
\end{figure}

\begin{table}[tb]
\centering
\begin{tabular}{c|c}
\hlinewd{1pt}
\cline{1-2}
  & Topic\\
\hline
1 & estate, real, construction\\
2 & workers, labor, wage\\
3 & loan, lending, credit \\
4 & agriculture, crop, corn\\
5 & oil, gas, drilling\\
6 & covid19, prepandemic, cases\\
\hlinewd{1pt}
\end{tabular}
\caption{Examples of extracted topics from FED dataset.}
\label{table:FEDTopics}
\end{table}

\subsubsection{Modality-aware Decoder}
We extend the modality-aware attention mechanism into MAT's decoder and introduce a novel target-modal MHA fused by feature-level attentions. 
This enables the model to discover the interactions between the target series and learnt patterns from two input data modalities toward decoding (forecasting) behavior of the target time series in the next timepoint(s).

Similar to \cite{vaswani2017attention}, at each step the model is auto-regressive, consuming the previously generated symbols as additional input to the decoder.
To prevent each position from attending to subsequent positions, a standard masked MHA with the residual connections following by layer normalization is used in the first sub-layer of the decoder. 
The masked MHA transforms the decoder's input into $h$ query $Q_{tar}$, key $K_{tar}$ and value $V_{tar}$ matrices, and then computes the scaled dot-product attention in each head to compute $Attn^{masked}$.

\begin{talign}
    Attn^{masked} &= \operatorname{softmax}\left(\frac{Q_{tar}K_{tar}^T}{\sqrt{d_{k}}}\right)V_{tar}
\end{talign}

\begin{talign}
    H_p &= Attn^{masked}(Q_{tar}W_{p}^Q, K_{tar}W_{p}^K, V_{tar}W_{p}^V)
\end{talign}

\begin{align}
\begin{split}
    Masked\ MHA = Concat(H_1,...,H_h)W_{tar}^O
\end{split}
\end{align}

\noindent where subscript $tar$ refers to the target time series, $h$ is the number of heads in the masked MHA and $W_{tar}^O$ are learnable parameters.

We developed a novel target-modal MHA in the second sub-layer which performs multi-head attention over the weighted outputs of the encoder after applying the feature-level attention in each modality stream. 
The target-modal MHA transforms the weighted output of the encoder into $h$ key $K_{mod_i}$ and value $V_{mod_i}$ matrices, and then computes the scaled dot-product attention with the target queries from the masked MHA sub-layer to compute $Attn_{mod_i}^{tar-mod}$.

\begin{align}
    Attn_{mod_i}^{tar-mod} = \operatorname{softmax}\left(\frac{Q_{tar}K_{mod_i}^T}{\sqrt{d_{k}}}\right)V_{mod_i}
\end{align}

\begin{align}
    H_p = Attn_{mod_i}^{tar-mod}(Q_{tar}W_{p}^Q, K_{mod_i}W_{p}^K, V_{mod_i}W_{p}^V)
\end{align}

\begin{align}
\begin{split}
    Target-modal MHA = Concat(H_1,...,H_h)W_{tar,mod_i}^O
\end{split}
\end{align}

Target-modal MHAs conduct the dependencies discovery between the generated target time series and each modality. 
Such proposed decoder’s structure provides more flexible and efficient approach to extract the most informative time steps where decoder’s input might have high correlation with different timesteps in each modality, and hence achieving more reliable and stable attention.

The proposed modality-aware structure also enables the model to deal
with different input sequences’ lengths from different modalities, supporting mismatched data sampling rates between modalities.
An approach of using a single lookback window can degrade the forecasting performance in applications with data modalities that are not step-wise synchronized with each other. To address this challenge, the proposed structure provides the flexibility to take sequences with different lengths from different modalities as input.

\subsubsection{Loss function} 
The proposed model is trained with the Mean squared error (MSE) as its loss function. Since the parameters of the model are updated using a mini-batch of training samples, the objective function is defined as below:
\begin{align}
\begin{split}
    J(\Theta_{MAT}) = \frac{1}{m} \Sigma_{k=1}^{m}\frac{1}{T\prime} \left\Vert MAT_\Theta(X_k)-Y_k \right\Vert^2_2
\end{split}
\end{align}

\noindent where $m$ is the batch size, ${T\prime}$ is the prediction horizon and $\Theta$ are the model’s parameters.

\begin{table*}[ht]
\centering
\begin{tabular}{l|cc|cc|cc}
\toprule
  \multirow{3}{*}{}  & \multicolumn{2}{c|}{1-month} & \multicolumn{2}{c|}{3-months} & \multicolumn{2}{c}{6-months} \\
   &  \multicolumn{2}{c|}{(LBW: 9 months)} & \multicolumn{2}{c|}{(LBW: 9 months)} & \multicolumn{2}{c}{(LBW: 9 months)}  \\
\cmidrule{2-7}
 Methods & MSE & MAE & MSE & MAE & MSE & MAE \\
\midrule
ElasticNet & 2.608 & 1.396 & 1.950 & 1.239 & 1.683 & 1.091 \\
\midrule
Transformer \cite{vaswani2017attention} &0.810 & 0.732 & 0.730 & 0.665 & 0.790 & 0.712\\
\midrule
Reformer \cite{kitaev2020reformer} &0.784 & 0.660 & 0.783 & 0.675 & \textbf{0.657} & \textbf{0.623}\\
\midrule
Autoformer \cite{wu2021autoformer} &0.936 &0.779  & 0.755 & 0.659 & 0.760 & 0.630\\
\midrule
Informer \cite{zhou2021informer} &0.779 & 0.712 & 0.868 & 0.742 & 0.709 & 0.653\\
\midrule
Modality-aware Transformer&\textbf{0.692} & \textbf{0.645} & \textbf{0.671} & \textbf{0.647} & 0.721 & 0.676\\
\bottomrule
\end{tabular}
\caption{Multi-step time series forecasting results with prediction length 1 month, 3 months and 6 months on FED dataset for 2 years interest rate maturity. MAE (lower is better), MSE (lower is better) are computed between the prediction results and target.}
\label{table:IR02}
\end{table*}

\begin{table*}[ht]
\centering
\begin{tabular}{l|cc|cc|cc}
\toprule
  \multirow{3}{*}{}  & \multicolumn{2}{c|}{1-month} & \multicolumn{2}{c|}{3-months} & \multicolumn{2}{c}{6-months} \\
   &  \multicolumn{2}{c|}{(LBW: 9 months)} & \multicolumn{2}{c|}{(LBW: 9 months)} & \multicolumn{2}{c}{(LBW: 9 months)}  \\
\cmidrule{2-7}
 Methods & MSE & MAE & MSE & MAE & MSE & MAE \\
\midrule
ElasticNet& 3.512 & 1.746 & 3.401 & 1.710 & 3.251 & 1.639 \\
\midrule
Transformer \cite{vaswani2017attention}& 0.544 & 0.579 & 0.809 & 0.723 & 1.030 & 0.838\\
\midrule
Reformer \cite{kitaev2020reformer} & 0.721 & 0.637 & 0.746 & 0.649 & 0.866 & 0.758\\
\midrule
Autoformer \cite{wu2021autoformer} & 0.638 & 0.650 & 0.968 & 0.773 & 0.824 & 0.728  \\
\midrule
Informer \cite{zhou2021informer} & 0.634 & 0.608 & 1.178 & 0.877 & 0.925 & 0.775\\
\midrule
Modality-aware Transformer& \textbf{0.494} & \textbf{0.519} & \textbf{0.613} & \textbf{0.594} & \textbf{0.752} & \textbf{0.696}\\
\bottomrule
\end{tabular}
\caption{Multi-step time series forecasting results with prediction length 1 month, 3 months and 6 months on FED dataset for 5 years interest rate maturity. MAE (lower is better), MSE (lower is better) are computed between the prediction results and target.}
\label{table:IR05}
\end{table*}

\begin{table*}[ht]
\centering
\begin{tabular}{l|cc|cc|cc}
\toprule
  \multirow{3}{*}{}  & \multicolumn{2}{c|}{1-month} & \multicolumn{2}{c|}{3-months} & \multicolumn{2}{c}{6-months} \\
   &  \multicolumn{2}{c|}{(LBW: 9 months)} & \multicolumn{2}{c|}{(LBW: 9 months)} & \multicolumn{2}{c}{(LBW: 9 months)}  \\
\cmidrule{2-7}
 Methods & MSE & MAE & MSE & MAE & MSE & MAE \\
\midrule
ElasticNet& 5.008 & 2.125 & 5.086 & 2.145 & 5.088 & 2.139 \\
\midrule
Transformer \cite{vaswani2017attention} & 0.872 & 0.764 & 0.951 & 0.817 & 1.256 & 0.962\\
\midrule
Reformer \cite{kitaev2020reformer} & 0.697 & 0.652 & 0.858 & 0.762 & 1.133 & 0.904\\
\midrule
Autoformer \cite{wu2021autoformer} & 0.674 & 0.636 & 0.983 & 0.891 & 1.217 & 0.996\\
\midrule
Informer \cite{zhou2021informer} & 0.751 & 0.663 & 1.429 & 0.991 & 1.447 & 1.041\\
\midrule
Modality-aware Transformer& \textbf{0.579} & \textbf{0.556} & \textbf{0.780} & \textbf{0.705} & \textbf{1.077} & \textbf{0.879}\\
\bottomrule
\end{tabular}
\vspace*{-0.2cm}
\caption{Multi-step time series forecasting results with prediction length 1 month, 3 months and 6 months on FED dataset for 10 years interest rate maturity. MAE (lower is better), MSE (lower is better) are computed between the prediction results and target.}
\label{table:IR10}
\end{table*}

\begin{table*}[ht]
\centering
\begin{tabular}{l|cc|cc|cc}
\toprule
  \multirow{3}{*}{}  & \multicolumn{2}{c|}{1-month} & \multicolumn{2}{c|}{3-months} & \multicolumn{2}{c}{6-months} \\
   &  \multicolumn{2}{c|}{(LBW: 9 months)} & \multicolumn{2}{c|}{(LBW: 9 months)} & \multicolumn{2}{c}{(LBW: 9 months)}  \\
\cmidrule{2-7}
 Methods & MSE & MAE & MSE & MAE & MSE & MAE \\
\midrule
ElasticNet& 5.798 & 2.283 & 5.906 & 2.310 & 5.956 & 2.327 \\
\midrule
Transformer \cite{vaswani2017attention} & 0.929 & 0.792 & 1.245 & 0.974 & 1.617 & 1.139\\
\midrule
Reformer \cite{kitaev2020reformer} & 0.930 & 0.812 & 1.119 & 0.918 & 1.488 & 1.096\\
\midrule
Autoformer \cite{wu2021autoformer} & 0.886 & 0.751 & 1.168 & 0.873 & 1.356 & 1.084\\
\midrule
Informer \cite{zhou2021informer} & 0.960 & 0.767 & 1.511 & 1.040 & 1.659 & 1.112\\
\midrule
Modality-aware Transformer& \textbf{0.741} & \textbf{0.665} & \textbf{0.901} & \textbf{0.799} & \textbf{1.262} & \textbf{0.994}\\
\bottomrule
\end{tabular}
\vspace*{-0.2cm}
\caption{Multi-step time series forecasting results with prediction length 1 month, 3 months and 6 months on FED dataset for 30 years interest rate maturity. MAE (lower is better), MSE (lower is better) are computed between the prediction results and target.}
\label{table:IR30}
\end{table*}

\section{Experimental Setup}
\subsection{Datasets}
We evaluate our proposed model through real-world applications. We collect U.S. Interest Rates (IR) data across various maturities (2 years, 5 years, 10 years and 30 years) along with text reports (FOMC statements and BeigeBook) from Federal Reserve's website\footnote{https://www.federalreserve.gov/} spanning a continuous period of twenty four years, from 1999 to 2022.

\vspace*{0.1cm}
\noindent\textbf{Learning Text Articles Representation}
Our textual module illustrated in Figure~\ref{fig:text} is used to represent each text report into a format comparable to time series that can be used along with the time series modality.
An input to the textual module contains a sequence of $n$ text reports $\{report_1,...,report_n \}$ collected in the last $T$ steps (there might be more than one text report per each timestep $t$). 
 We start modeling from the level of sentences by disaggregating text articles into sentences. 
 Since the number of sentences are large and varied per each timestep, the sentences are aggregated to the topic level. 
To this end, the topic model \cite{grootendorst2022bertopic} is used to extract sentences embeddings with a pre-trained language model and cluster them to generate the corresponding topic representations. Then, sentences from the text report are assigned to the generated clusters (topics). Some of the extracted topics from FED dataset are shown in Table~\ref{table:FEDTopics}.
Finally, the textual module exploits the domain-adapted financial BERT model \cite{araci2019finbert} to compute the sentiment score of each sentence and aggregates each text report into topic level by averaging the sentiment scores per topic. 
We apply our text module to every text report collected at timestep $t$, and convert it into a sequence of $k$ representative topics $\{s_1,...,s_k \}$ where $s_j$ corresponds to the average sentiment score per topic $j$ ($t$ is omitted for simplicity).

We further collect various indices/timeseries of the economy such as unemployment rate, retail sale, housing price, oil price (released by FRED \footnote{https://fred.stlouisfed.org/}) during the same time period, which we call FRED time series data. Along with the U.S. interest rates, we down sample these time series into the monthly basis through averaging the values. 

Each sample point consists of two data modalities: (i) a sequence of interest rates and FRED time series data; (ii) 
a sequence of text documents from all FED reports in the same time period.
A model is trained to forecast the future of U.S. interest rates for the predication length 1-month, 3-months and 6-months ahead in a monthly basis.
We split the data into 70\%, 15\%, 15\% for training, validation and testing. The input of the dataset is zero-mean normalized.

\subsection{Implementation Details} 
The model is trained using Adam optimizer~\cite{kingma2014adam} with an initial learning rate of $10^{- 4}$ and a batch size of 16. The training process is early stopped within 10 epochs. Consistent with transformer-based time series forecasting approaches \cite{zhou2021informer,kitaev2020reformer}, we configure the model with 8 heads for multi-head attention and an output dimension of 512 for each head. Our implementation is based on PyTorch and training is performed on a machine equipped with a 16GB GPU.

\paragraph{Compared models:}
We select four transformer-based time series forecasting methods for comparison, including Transformer \cite{vaswani2017attention}, Informer \cite{zhou2021informer}, Reformer \cite{kitaev2020reformer}, Autoformer \cite{wu2021autoformer} 
and one classical time series forecasting method ElasticNet~\cite{trevor2009elements}. To better evaluate our proposed model, we use the same set of features and parameters for all methods in the comparison.
All models are built with two encoder layers and one decoder layer for the sake of the fair comparison.
Although our proposed model is able to deal with input data sequences of different lengths, the same sequence lengths of two modalities are used in all experiments to have its fair comparison with other methods. 
All experiments are performed using the same training, validation and testing splits.
We used publicly available source codes for the transformer-based time series forecasting baselines.

\textbf{Metrics}: Two commonly-used quantitative measures,
$MSE = \frac{1}{n} \sum\limits_{t=1}^{n} (y_t - \hat{y_t})^2$
and 
$MAE = \frac{1}{n} \sum\limits_{t=1}^{n} |y_t - \hat{y_t}|$ 

are used for evaluation, with $y_t$ and $\hat{y_t}$ are respectively the groundtruth and predicted signals at timepoint $t$, and $n$ is the total number of timepoints in prediction horizon. 
Lower values of MSE and MAE indicate better prediction accuracy.

\section{Experimental Results}
Extensive experiments were performed to compare MAT with current state-of-the-art transformer-based models and one classical time series forecasting model for interest rate forecasting.
Tables~\ref{table:IR02}, ~\ref{table:IR05}, ~\ref{table:IR10}, and ~\ref{table:IR30} summarize the evaluation results of all comparison methods on the FED dataset for different interest rate maturities of 2 years, 5 years, 10 years, and 30 years, respectively. 
To compare performances under different prediction horizons, we fix the input length to 9 months and evaluate models with prediction lengths: 1-month, 3-months and 6-months. 
Under the same setting, each method forecasts the U.S interest rate as a single variable
using both FRED data (timeseries modality) and FED reports (text modality) as input. The best results are highlighted in boldface.

As shown in Tables~\ref{table:IR02}, ~\ref{table:IR05}, ~\ref{table:IR10}, and ~\ref{table:IR30}, our proposed model significantly improves the inference performance across all interest rate maturities especially for the long-term maturities such as 10years and 30years that are more challenging for modeling. The results clearly demonstrate the effectiveness of the proposed modality-aware structure in multimodal time series forecasting compared to all competing methods.

Also, we observed that Reformer performs slightly better on the short-term maturity 2years (6-months prediction horizon and 2years interest rate setting), and our method surpasses on long-term maturities such as 5years, 10years, and 30years. 
This is possibly because of forecasting short-term interest rate such as 2years is simpler for modeling and all methods achieve similar performance.

\subsection{Comparison against a statistical pipeline}
We further evaluated our approach against the Sentometrics pipeline \cite{ardia2019questioning}, a non-deep-learning model which adopts a linear regression as a forecasting model along with its feature engineering based on text lexicons.
Sentometrics has its own pipeline that extracts sentiment scores from the text corpus using various lexicons and beta weighting functions to aggregate input sequences into vectors so that a linear regression can be applied. We adopt its whole pipeline as provided in the Sentometrics package \footnote{https://cran.r-project.org/web/packages/sentometrics/}. It is worth mentioning that our approach is competitive and achieves better performance especially in forecasting long-term interest rates such as 10years and 30years. 
This demonstrates the effectiveness of the proposed Modality-aware Transformer in exploiting information within each modality as well as across modalities to better forecasting future behavior of the target time series.

\vspace*{-0.2cm}
\section{Conclusion}
In this paper, we presented the Modality-aware Transformer (MAT) as a novel solution for addressing the challenges associated with multimodal time series forecasting. MAT effectively harnesses information from individual modalities while also facilitating cross-modality learning through its innovative modality-aware structure, resulting in more accurate forecasting performance. Its notable strength lies in its flexibility, particularly in aligning data modalities at the timestamp level, which is instrumental in achieving superior forecasting performance compared to existing state-of-the-art methods demonstrated through extensive experiments on the real-world datasets. Our work highlights the potential of MAT in advancing the field of multimodal time series forecasting, representing as a promising model in domains where multimodal data sources are prevalent.

\bibliographystyle{siam}
\bibliography{ref}
\vspace*{-0.5cm}

\end{document}